\documentclass[a4paper,fleqn]{cas-sc}

\usepackage[authoryear,longnamesfirst]{natbib}

\def\tsc#1{\csdef{#1}{\textsc{\lowercase{#1}}\xspace}}
\tsc{WGM}
\tsc{QE}

\begin{document}
\let\WriteBookmarks\relax
\def\floatpagepagefraction{1}
\def\textpagefraction{.001}

\shorttitle{Hyperbox Mixture Regression}    


\title [mode = title]{Hyperbox Mixture Regression for Process Performance Prediction in Antibody Production}  



%

\author[1]{Ali Nik-Khorasani}
\ead{al.nikkhorasani@mail.um.ac.ir}

\author[2]{Thanh Tung Khuat}
\ead{thanhtung.khuat@uts.edu.au}
\cormark[1]

\author[2]{Bogdan Gabrys}
\ead{bogdan.gabrys@uts.edu.au}

\credit{}

\affiliation[1]{organization={Department of Electrical Engineering},
            addressline={Ferdowsi University of Mashhad}, 
            city={Mashhad},
            country={Iran}}

\credit{}

\affiliation[2]{organization={Complex Adaptive Systems Laboratory, Data Science Institute},
            addressline={University of Technology Sydney},
            city={Sydney},
            country={Australia}}

\cortext[1]{Corresponding author}


\begin{abstract}
This paper addresses the challenges of predicting bioprocess performance, particularly in monoclonal antibody (mAb) production, where conventional statistical methods often fall short due to time-series data's complexity and high dimensionality. We propose a novel Hyperbox Mixture Regression (HMR) model which employs hyperbox-based input space partitioning to enhance predictive accuracy while managing uncertainty inherent in bioprocess data. The HMR model is designed to dynamically generate hyperboxes for input samples in a single-pass process, thereby improving learning speed and reducing computational complexity. Our experimental study utilizes a dataset that contains 106 bioreactors. This study evaluates the model's performance in predicting critical quality attributes in monoclonal antibody manufacturing over a 15-day cultivation period. The results demonstrate that the HMR model outperforms comparable approximators in accuracy and learning speed and maintains interpretability and robustness under uncertain conditions. These findings underscore the potential of HMR as a powerful tool for enhancing predictive analytics in bioprocessing applications.
\end{abstract}


\begin{highlights}
\item Introduces Hyperbox Mixture Regression for effective bioprocess performance prediction.
\item Develops fast learning method, enhancing efficiency in high-dimensional data.
\item Incorporates hyperbox fuzzy sets for improved model transparency and understanding.
\item Tackles bioprocessing challenges using advanced machine learning techniques effectively.
\item Accurately forecasts monoclonal antibody production indicators up to two days.
\end{highlights}


\maketitle

\section{Introduction}\label{}
Regression models have found widespread application in various fields, including robot controllers \citep{RN2}, motion prediction \citep{RN3}, and time series forecasting \citep{lega10,ruga11,RN1}. Predicting bioprocess performance presents a complex multivariate time-series challenge that conventional statistical methods often struggle to address \citep{RN19}. While numerous types of research have focused on data pre-processing techniques—such as imputation, visualization, and feature selection—choosing a suitable predictive model remains a critical hurdle~\citep{khba24}. This paper aims to tackle these challenges by developing a machine-learning model specifically designed for bioprocess performance prediction.

The growing complexity of time-series data has led to an increasing reliance on machine learning (ML) techniques to overcome the limitations of traditional statistical methods. These conventional methods often struggle with the inherent correlations in time-series observations, resulting in potential inaccuracies in predictions \citep{gangadharan2019metaheuristic}. In contrast, ML methods have gained popularity for extracting essential information from time-series data, providing more robust and accurate insights \citep{lim2023opportunities}. This trend underscores a significant shift towards leveraging ML as a powerful tool for addressing the challenges inherent in time-series analysis.

Recent studies have demonstrated the effectiveness of ML algorithms in predicting critical quality attributes (CQAs) and process outcomes. For instance, \citet{khba24} highlights the growing applications of ML in biopharmaceuticals, emphasizing its role in real-time monitoring and optimization of both upstream and downstream processes. By leveraging large datasets generated from production, ML models can identify patterns and relationships that are not easily discernible through conventional statistical methods. In addition, \citet{waight2023machine} argued that identifying favorable biophysical properties is essential in the preclinical development of protein therapeutics, but predicting these properties remains challenging. They introduced an automated machine learning workflow that analyzes computationally derived features to build predictive models for key developability factors like hydrophobicity and poly-specificity in IgG molecules. More studies have been reviewed in \citet{lim2023opportunities, khba24}. This approach addresses some of the challenges in preclinical development, where predicting favorable biophysical properties is crucial yet difficult. 

One of ML's key advantages is its ability to handle bioprocess data's high dimensionality and complexity. ML algorithms, such as random forests, support vector machines, and deep learning models, have been applied to predict CQAs and key performance indicators (KPIs) in monoclonal antibody (mAb) production. For instance, a recent study reports a deep learning-based 2D-convolutional neural network (2D-CNN) designed to predict various downstream processing attributes, including Protein A mAb elute concentration and aggregate percentages, from routinely collected process data~\citep{alam2024deep}. According to \citet{alam2024deep}, their model outperformed existing approaches, achieving a mean percentage deviation of less than $3\%$ in experimental validation. In another study, Lai employed machine learning to predict therapeutic antibody aggregation rates and viscosity at high concentrations ($150^{mg/ml}$), focusing on preclinical and clinical-stage antibodies. \citet{lai2022machine} employed a k-nearest neighbors regression model and achieved a high correlation for predicting aggregation rates using features derived from molecular dynamics simulations. Moreover,  \citet{schmitt2023predictive} employed an Artificial Neural Network (ANN) to predict mAb viscosity. High concentrations of mAb solutions can increase viscosity, affecting protein purification and administration. They utilized an ANN and combined experimental factors and simulated data to predict and model the viscosity of mAbs. 

Additionally, new research by \citet{makowski2024reduction} has shown the use of a transparent machine-learning model for predicting antibody (IgG1) variants with low viscosity based on the sequences of their variable (Fv) regions. This model not only identifies antibodies at risk for high viscosity with relatively high accuracy but also enables the design of mutations that reduce antibody viscosity, confirmed experimentally. According to Makowski et al., their model demonstrates high accuracy and exhibits excellent generalization \citep{makowski2024reduction}. These advancements underscore the growing role of ML and deep learning in enhancing the efficiency and quality of mAb production processes.

In terms of predicting mAb stability, recent work has focused on chemical modifications, such as methionine oxidation, that can impair antibody potency. A study developed a highly predictive in silico model for methionine oxidation by extracting features from mAb sequences, structures, and dynamics, utilizing random forests to identify crucial predictive features \citep{sankar2018prediction}. This work emphasizes the potential for computational tools to complement experimental methods in therapeutic antibody discovery.

However, despite the potential benefits, challenges remain in the widespread adoption of ML in bioprocessing. Issues such as limited samples, high-dimensional data, data quality, model interpretability, and robust validation protocols must be addressed to ensure reliable application in industrial settings \citep{RN19, gangadharan2019metaheuristic,khba24}. In addition, the model should be capable of making inferences under uncertain conditions and, ideally, provide explanations for its predicted outcomes \citep{lim2023opportunities}. While neuro-fuzzy regression models \citep{RN4, de2020fuzzy} are effective in managing uncertainty, they face challenges in the high-dimensional data space \citep{RN7}. Conversely, regression models such as Radial basis functions (RBF) and ANN, which excel in handling high-dimensional data spaces \citep{sung2004gaussian}, struggle to address the inherent uncertainty in the problem. It motivates us to develop a neuro-fuzzy system that handles uncertainty within high-dimensional data spaces for bioprocess performance prediction.

RBF is a regression and classification model that can be employed when the relationship between variables is unknown. Key advantages of RBF models are guaranteed learning algorithm through linear least squares optimization and efficiency in dealing with high dimensional data \citep{walczak1996radial}. This model consists of an unsupervised clustering framework that helps partition the input feature space, then estimate the target signal using least squares optimization \citep{tagliaferri2001fuzzy, walczak1996radial}. However, a significant challenge for RBF networks is determining the optimal number and distribution of nodes in the hidden layer \citep{walczak1996radial}. 

Another class of powerful neuro-fuzzy machine learning models, which are of particular interest to us, are based on hyperbox fuzzy sets originally introduced by Simpson in the 1990s \cite{simpson1993fuzzy} and then later improved, extended, and generalized by Gabrys \cite{RN17,ga02,ga02a,ga04} as well as a large number of other researchers \cite{khru21}. In its original paper, Simpson introduced the Fuzzy Min-Max (FMM) algorithm as an unsupervised clustering method for pattern clustering \citep{simpson1993fuzzy}. FMM is a neuro-fuzzy algorithm that integrates fuzzy inference systems and adaptive neural networks. Employing a fuzzy inference system facilitates the creation of a neuro-fuzzy system capable of handling uncertainty. Additionally, using an adaptive neural network structure allows one to use learning approaches to find optimal parameters \citep{RN5}. Furthermore, having the ability to extract fuzzy if-then rules from the network architecture means that it is no longer a black box model. As mentioned earlier, unsupervised clustering methods can be utilized for partitioning of an input space in neural network based regression (e.g. RBF) models and hence there have also been some examples of FMM-based regression models.

Simpson and Jahns introduced an FMM-based framework for function approximation \citep{simpson1993fuzzyb}. In their approach, the authors utilized the FMM clustering method to partition the input feature space and used the hyperbox fuzzy sets representing clusters and the associated trapezoidal fuzzy membership functions as basis functions to estimate the target output. Similarly to the RBF networks, the output was a weighted combination of hyperbox fuzzy sets membership values \citep{simpson1993fuzzyb}.

In another study, Tagliaferri developed an innovative FMM-based model for function approximation, enhancing the FMM clustering algorithm for better feature space partitioning \citep{tagliaferri2001fuzzy}. Tagliaferri asserted that batch learning algorithms, which partition the feature space using the entire dataset, help eliminate the dependence on the order of data presentation. According to Tagliaferri, this adjustment significantly enhances the model's performance \citep{tagliaferri2001fuzzy}.

Additionally, Brouwer proposed a novel automatic learning algorithm designed for the FMM-based function approximation model \citep{brouwer2005automatic}. Given that the loss function of the FMM is not differentiable and thus incompatible with gradient descent optimization, Brouwer implemented a helper neural network to approximate the loss function. This network allowed the application of the gradient descent algorithm to adjust the network parameters. Brouwer's new learning framework demonstrated the capability to achieve superior optimal values compared to conventional training algorithms \citep{brouwer2005automatic}. 

All of these RBF-like models involve clustering-based input feature space partitioning, which helps manage high-dimensional input data and overcome a curse of dimensionality problem quite common in other neuro-fuzzy regression methods like ANFIS \cite{RN4}. 

Despite some initial interest, successees and many attractive features there have been far fewer hyperbox regression models than their classification/clustering counterparts \cite{khru21}. In this paper, we therefore introduce a novel neuro-fuzzy structure and learning procedure called Hyperbox Mixture Regression (HMR) that utilizes the hyperbox input space partitioning to overcome the curse of dimensionality problem while taking full advantage of its universal approximator capabilities and model transparency. The proposed method employs the hyperbox idea in constructing the basis functions in the first layer. It is worth noting that the new HMR structure is more straightforward than conventional neuro-fuzzy structures like ANFIS \cite{RN4} and has a lower computational complexity to produce the output. 

A hyperbox is a convex n-dimensional box in the feature space that assigns a full membership value to patterns within it \citep{RN13, khru21} and is defined by its maximum and minimum points \citep{RN13}. The utilization of hyperboxes for the input space partitioning offers notable advantages. A key benefit is their ability to learn from an input sample in a single-pass process, leading to a significant boost in the learning speed of the system \citep{RN14, RN17, ga02}. Moreover, the learning framework of hyperboxes is free from limitations such as being trapped in a local minimum or divergence due to the existence of outliers. Additionally, employing hyperboxes enables the system to identify essential basis functions, reducing overall complexity, particularly in high dimensions.

The ability of the HMR model to infer and explain under uncertain conditions using the generated hyperbox fuzzy sets is crucial in industrial fields such as mAb production, where measurement noise is frequently associated with process parameters. This paper will apply the proposed HMR model to predict the performance of mAb production processes based on the critical process parameters used as input features \citep{RN19}. The empirical dataset encompasses information from 106 bioreactors, capturing biological parameters over 15 cell culture days. The model's primary objective is to predict values of key performance indicators, such as Viable Cell Density (VCD) and mAb concentration, for antibody production processes within the subsequent two days. Consequently, the development involves the creation of two distinct predictive models: one for forecasting antibody production one day ahead and another for predicting the production two days ahead. The main contributions of this paper can be summarized as follows:

\begin{enumerate}
    \item Introducing a new neuro-fuzzy model structure for bioprocess performance prediction to address the limited samples, high-dimensional data space, and transparency limitations under uncertain conditions.
    \item Introducing a novel learning procedure that learns in a single input data pass process and generates essential basis functions for regression, effectively increasing the learning rate and reducing system complexity in high dimensional problems.
    \item Employing a dynamically weighted combination of local linear regressors, associated with each hyperbox, in order to increase accuracy and decrease network complexity, especially in non-linear problems.
    \item Introducing a normalization layer in the proposed structure to reduce the risk of numerical instability in the next layers.
    \item Forecasting the key performance indicators of antibody production processes over the next two cell culture days.
\end{enumerate}

The rest of the paper is organized as follows. The next section describes in detail and illustrates the proposed model structure and training algorithm. In the third section, comprehensive experiments evaluate the proposed model in different scenarios. Finally, section four concludes the paper and highlights the key contributions.

\section{Proposed Method}
In this section, the proposed model is described in detail. The section is divided into two parts representing the proposed model structure and learning procedure.
\begin{figure}[t]
    \centering
    \includegraphics[scale=0.5]{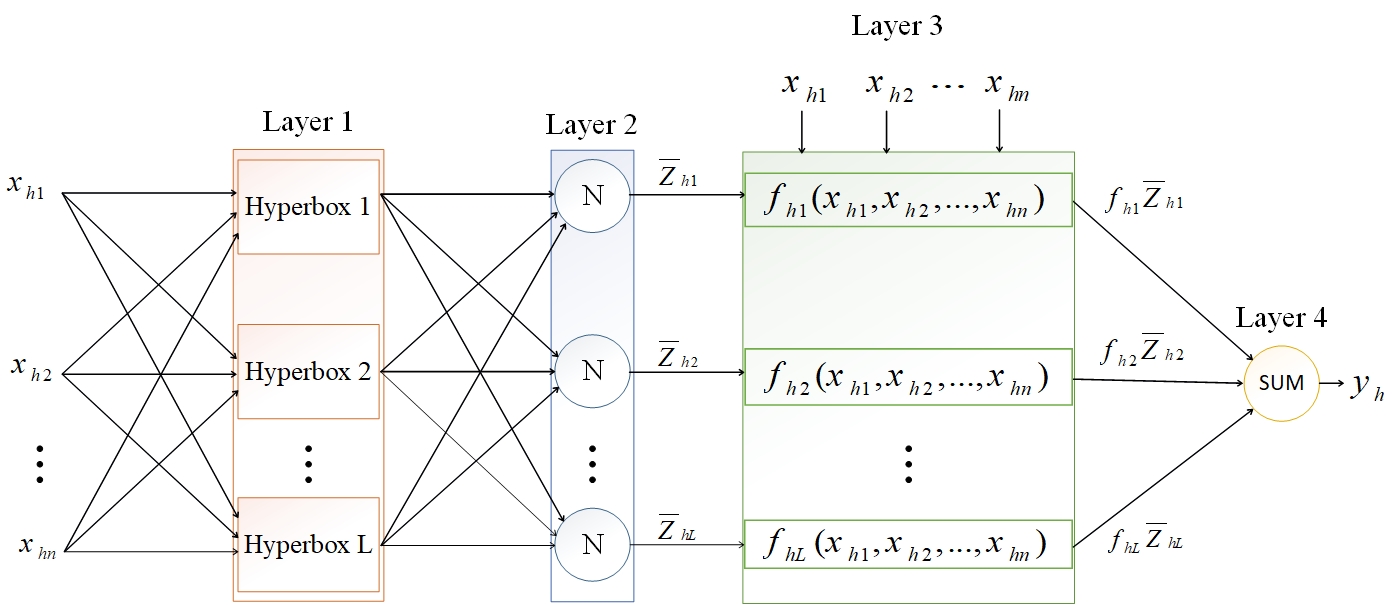}
    \caption{\label{Fig1}Proposed HMR structure with $L$ radial functions}
    \label{fig:enter-label}
\end{figure}
\subsection{Hyperbox Mixture Regression Structure}
Figure \ref{Fig1} illustrates the proposed HMR structure. According to the Fig \ref{Fig1}, the HMR consists of four layers. In the first layer, each node represents a fuzzy hyperbox and computes membership values for the input sample. The number of nodes in the first layer is determined dynamically during the learning stage. The second layer normalizes the computed membership values and reduces the risk of numerical instability that can arise in the next layers. In the third layer, we utilize a linear regressor for each hyperbox from the previous layer. The final network output is computed using the sum of the weighted, local (i.e. associated with each hyperbox) linear regressors. The following equations compute the system's output based on this structure.
\begin{equation*}
\text{Layer 1: } u_{HB_l} \text{ for } l=1,2, \ldots, L
\end{equation*}
\begin{equation}\label{eq1}
 u_{HB_l}(X_h)=\min_{i=1,\dots,n}(\min([1-g(x_{hi}-w_{li},\lambda_i)],[1-g(v_{li}-x_{hi},\lambda_i)]))
\end{equation}
where $i=1,2,\dots,n$ shows the feature index, $L$ is the number of hyperboxes, $u_{HB_l}$ represents the membership value of the $l^{th}$ node in the interval $[0,1]$, $v_{li}$ and $w_{li}$ show minimum and maximum of the hyperbox points in the feature space, $X_h=(x_{h1}, x_{h2}, ..., x_{hn})$ represents the $h^{th}$ input sample, $n$ shows the number of input variables, and $g(r,\lambda_i)$ can be computed using the following Eqs.~\eqref{eq2}:
\begin{equation}\label{eq2}
g(r,\lambda_i)=\begin{cases}
1 & \textrm{if $r\lambda_i>1$}\\
r\lambda_i & \textrm{if $0\le r\lambda_i \le 1$}\\
0& \textrm{if $r\lambda_i<0$}
\end{cases}
\end{equation} 

where $\lambda_i$ is a sensitivity coefficient for the hyperbox. The second layer is the normalization layer, which normalizes computed membership values.
\begin{equation}\label{eq5}
\text{Layer 2: } \bar{Z}_{hl}=\frac{u_{HB_l}(X_h)}{\sum_{j=1}^{L}u_{HB_j}(X_h)} \text{ for } l=1,2, \ldots, L
\end{equation}
where $\bar{Z}_{hl}$ indicates the normalized membership value for the $h^{th}$ sample. 

In the third layer, the $f_{hl}$ function for each hyperbox is computed using Eq.~\eqref{eq6}:
\begin{equation}\label{eq6}
\text{Layer 3: } f_{hl}=\sum_{i=1}^{n}d_{li}x_{hi}+r_{l} \text{ for } l=1,2,\ldots,L 
\end{equation}
where the $d_{li}$ and $r_{l}$ are the local linear regression function parameters and will be optimized during the learning stage. Please not that $f_{hl}$ can take different forms from $f_{hl}=r_{l}$ in which case the output would be a sum of the weighted fuzzy membership values to locally trainable highly nonlinear functions. These other forms of $f_{hl}$ are however outside of the scope of the current paper. 

The output of the system can be obtained using Eq.~\eqref{eq7}:
\begin{equation}\label{eq7}
\text{Layer 4: } y_h=\sum_{l=1}^{L}f_{hl}\bar{Z}_{hl}
\end{equation}
where $y_h$ is the system's output, the HMR parameters must be optimized using a learning procedure to ensure accurate estimation. The next section introduces a new learning procedure for the proposed HMR structure. 

\subsection{Hyperbox Mixture Regression Learning Procedure}
This section describes a novel learning procedure to optimize the HMR parameters. The learning algorithm requires a single input data pass process for feature space partitioning, significantly increasing the learning speed. 

The proposed HMR learning algorithm comprises two stages: 
Hyperbox Min-Max clustering and the Least Squares Optimization (LSO). 

Hyperbox min-max clustering involves creating hyperboxes and producing the first layer of the model. In the second stage, the least squares optimization is used to find the local $f_{hl}$ regressors parameters. The rest of this section describes the proposed learning procedure in detail.	

\subsubsection{Hyperbox Min-Max Clustering}
As mentioned before, the hyperboxes are created and form the first layer of the network. This stage involves:
\begin{enumerate}
    \item Computing membership values
    \item Selecting the top-$K$ winning hyperboxes
    \item Checking the expandability
    \item Expanding the selected winning hyperbox
\end{enumerate}
Consider an input sample; Eq.~\eqref{eq1} computes the sample's membership values and finds the winning hyperbox with the highest membership value. Then, the algorithm expands the winning hyperbox to contain the input sample. If the winning hyperbox does not meet the expansion criterion, the algorithm checks the next winner until the $K^{th}$ winner. If all $K$ winning hyperboxes do not meet the expansion criterion, the algorithm generates a new hyperbox containing the input sample.

The following equation indicates the expansion criterion for each dimension $i$ of the $l^{th}$ hyperbox:
\begin{equation}\label{eq8}
    \theta \geq \max(w_{li},x_{hi})-\min(v_{li},x_{hi}),
\end{equation}
where $\theta$ is the expansion coefficient ranging between $[0,1]$. It will be expanded if the expansion criterion is satisfied for $60\%$ of the dimensions within the winning hyperbox \citep{kumar2019improved}. This criterion leads to generating fewer hyperboxes, and as a result, the system will have lower complexity \citep{kumar2019improved}. The following equations illustrate the expansion formula:

\begin{equation}\label{eq9}
v_{li}^{new}=\min(v_{li}^{old},x_{hi})
\end{equation}
\begin{equation}\label{eq10}
w_{li}^{new}=\max(w_{li}^{old},x_{hi})
\end{equation}

This algorithm creates the necessary hyperboxes for the first layer. Therefore, this stage generates the required basis functions for regression. The number of created basis functions can be adjusted using the expansion coefficient $\theta$. In other words, when we choose a large value of $\theta$, the number of created basis functions decreases. On the other hand, a smaller value of $\theta$ can lead to the system creating more basis functions. It is essential to highlight that selecting a small value for $\theta$ is preferable when dealing with a highly non-linear target. Conversely, opting for a larger value of $\theta$ is suitable for less complex functions.

 \subsubsection{Least Squares Optimization}
This stage aims to optimize the local regressors parameters. These parameters can be optimized using the LSO algorithm. Consider the following equation:
\begin{equation}\label{eq11}
y_h=\sum_{l=1}^{L}f_{hl}\bar{Z}_{hl}=\sum_{l=1}^{L}(\sum_{i=1}^{n}(\bar{Z}_{hl}d_{li}x_{hi})+\bar{Z}_{hl}r_{l})
\end{equation}
Eq.~\eqref{eq11} can be rewritten as the following equation:
\begin{equation}\label{eq12}
Y=AD
\end{equation}
where $D$ is an $S \times 1$ vector that contains the local regressors parameters. $A$ represents an $N \times S$ matrix described as in Eq.~\eqref{eq13}. In addition, $Y$ shows an $N \times 1$ vector that contains the output targets. Furthermore, $S$ and $N$ indicate the total number of local regressors parameters and input samples.
\begin{equation}\label{eq13}
\text{\ensuremath{{\scriptstyle A}=}}{\scriptstyle \left[\begin{array}{cccccc}
\ensuremath{{\scriptstyle a_{11}}} & \ensuremath{{\scriptstyle a_{12}}} & {\scriptstyle \cdots} & {\scriptstyle a_{1L}}\\
\ensuremath{{\scriptstyle a_{21}}} & \ensuremath{{\scriptstyle a_{22}}} & {\scriptstyle \cdots} & {\scriptstyle a_{2L}}\\
{\scriptstyle \vdots} & {\scriptstyle \vdots} & {\scriptstyle \ddots} & {\scriptstyle \vdots}  \\
\ensuremath{{\scriptstyle a_{N1}}} & \ensuremath{{\scriptstyle a_{N2}}} & {\scriptstyle \cdots} & {\scriptstyle a_{NL}}
\end{array}\right]} 
\end{equation}
\begin{equation}
a_{hl} = \left[ \bar{Z}_{hl}x_{h1}, \bar{Z}_{hl}x_{h2}, \ldots, \bar{Z}_{hl}x_{hn}, \bar{Z}_{hl}\right]
\end{equation}
\begin{equation}
D = \left[ d_{11}, \ldots, d_{1n}, r_1, \ldots, d_{L1}, \ldots, d_{Ln}, r_L\right]^{T}
\end{equation}
\begin{equation}\label{eq15} 
Y = \left[y_1, y_2, \ldots, y_N\right]^{T}
\end{equation}
Optimal values for $D$ can be computed by optimizing the following square loss function: 
\begin{equation}\label{eq16}
\mathcal{L}_s=||AD-Y||^2
\end{equation}
Therefore, optimal values for $D$ can be obtained using Eq.~\eqref{eq17}:
\begin{equation}\label{eq17}
D=(A^TA)^{-1}A^TY
\end{equation}
where $A^T$ shows the transpose of the matrix $A$, and $(A^TA)^{-1}$ indicates the pseudo-inverse of $A$. 
\section{Experimental Results}
The experiments in this paper were performed on a bioprocess dataset given in \citep{RN19} to evaluate the efficacy of the learning models. The dataset comprises data from 106 cultures and encompasses 23 critical process parameters per culture. These parameters include mAb, Viable cell density (VCD), Elapsed Culture Time (ECT), Elapsed Generation Number (EGN), Total cell density (TCD), potassium concentration ($K^+$), etc. The dataset spans a collection period of 15 days for each culture. The objective of the dataset is to predict the values of mAb and VCD for the upcoming two days. However, the dataset presents a challenge with 3074 missing values filled using the missing value imputation method presented in \citep{RN19}. 

The proposed model addresses the curse of dimensionality associated with some of the popular fuzzy-based models like ANFIS \cite{RN4}. Therefore, the experiments are designed to evaluate the model in both high-dimensional and low-dimensional scenarios and compare the proposed HMR model with the ANFIS Hybrid Learning (HL) and Fuzzy Neural Network (FNN) Back-Propagation (BP) algorithms \citep{ANFIS2024robust}. The comparison focuses on model complexity, learning time, and model performance. Predictive models were developed to forecast future values of mAb concentration and VCD based on the current day's process parameter inputs.

The generalization of predictive algorithms is assessed using the 5-fold cross-validation method. Therefore, 106 cultures were divided into five folds; four were used for training, and the remaining fold was used for testing. This training and testing process was repeated for all 5-folds, and the reported results are the mean of five experiments corresponding to each testing fold. This approach evaluates how well the predictive models generalize in predicting the $mAb$ concentration and $VCD$ values for the cultures the trained model has not previously encountered. Consider the following vector:
\begin{equation}
    C_i(t)=[ECT_i(t), VCD_i(t),\ldots, Glutamine_i(t)]_{1\times23},	
\end{equation}
where $C_i(t)$ indicates the $i^{th}$culture that contains $23$ process parameters (features) in the respective $t^{th}$ day ($t$ is a discrete variable). Consider $C_1(t)$ as an example, for VCD prediction, $C_1(t)$ corresponds to the first culture with $23$ input features on the respective $t^{th}$ day, where $1\le t\le 15$. The $C_1(t)$ target is $VCD_1(t+1)$, representing the $VCD$ value at the day $t + 1$ for the first culture. Additionally, for $mAb$ values, the target follows a similar pattern; for instance, the target for $C_i(t)$ would be $mAb_i(t+1)$.

In the rest of this section, comprehensive experiments are conducted to evaluate and compare the proposed model with other competing models. The first experiment (Subsection \ref{part_1}) evaluates models' performance for one-day-ahead mAb and VCD prediction in high-dimension scenarios, including all 23 input features. The second experiment (Subsection \ref{part_2}) uses a feature selection algorithm to reduce the input feature space. The third experiment (Subsection \ref{part_3}) compares alternative models' accuracy, complexity, and learning rate for one-day-ahead mAb and VCD prediction in low dimensional scenarios after feature selection. In the last part (Subsection \ref{part_4}), comprehensive experiments are conducted to evaluate the HMR model for one-day-ahead and two-day-ahead mAb and VCD prediction. In all following experiments, the $\lambda_i$ is set to the value of $1$.
\begin{table}
\centering
\caption{\label{tab1}The testing RMSE and standard deviation scores of different predictive models using all 23 input features over 5-fold cross-validation}%
\begin{tabular}{c c c c c}
    \toprule
         Model & \multicolumn{2}{c}{${VCD}_i(t+1)$} & \multicolumn{2}{c}{${mAb}_i(t+1)$}
         \tabularnewline
    \midrule 
         & Training RMSE & Testing RMSE & Training RMSE & Testing RMSE 
        
        \tabularnewline
        ANFIS HL& $0.3319 \pm 0.0074$ & $0.3321 \pm 0.0494$ & $0.2553 \pm 0.0034$ & $0.2815 \pm 0.0382$
        \tabularnewline
        FNN BP& $0.1711  \pm 0.0048$ & $0.1727  \pm 0.0285$ & $0.1598  \pm 0.0009$ & $0.1606  \pm 0.0121$
        \tabularnewline
        HMR & $ 0.0283 \pm 0.0010 $ & $ 0.0537\pm 0.0238$ & $ 0.0296 \pm 0.0017 $ & $ 0.0516 \pm 0.0110 $
        \tabularnewline
        \bottomrule
    \end{tabular}
\end{table}
\subsection{High dimensional scenario}\label{part_1}
We trained the predictive models using all 23 input features in the first part. The 5-fold cross-validation method was used to assess the performance of the trained models. Therefore, for each of the five experiments, we have:
\begin{itemize}
    \item Training dataset: all $C_i(t)$  where $i$ are cultures within the four training folds, and $1\le t\le 14$
    \item Testing dataset: all $C_i(t)$  where $i$ are cultures within the testing fold, and $1\le t\le 14$
    \item The targets of datasets are $VCD_i(t+1)$  and $mAb_i(t+1)$.
\end{itemize}

Table \ref{tab1} illustrates the results of predictive models for all input variables using 5-fold cross-validation. Due to the curse of dimensionality, the original ANFIS models pose a significant computational burden when utilizing 23 input variables. For instance, if the ANFIS creates only two membership functions for each input feature, the grid partitioning leads to the generation of $2^{23}$ rules, leading to challenges in optimization. This limitation necessitates selecting only one membership for each input variable, consequently diminishing the performance of the original ANFIS models. In contrast, HMR dynamically generates basis functions without such constraints. Therefore, HMR significantly outperforms the ANFIS HL algorithm and FNN BP in the high-dimensional scenario. 

\subsection{Feature Selection}\label{part_2}
In the following experiments, we conducted a feature selection procedure to reduce the input feature space. The feature selection algorithm utilizes 5-fold cross-validation to select the most important input features within the training data. For this purpose, we split the training data into five folds: four folds are utilized for feature selection, and the other is employed for validation purposes. Then, the top-k features with the highest correlation that contribute to reaching the lowest RMSE scores on the validation folds are selected. After feature selection, the selected features are evaluated using the testing dataset. The HMR model was chosen as the predictor in the feature selection process due to its learning speed and dynamic basis function generation. The experiment reordered the input variables based on their correlation scores with the target in the training fold. It is assumed that input variables with a higher correlation with the target might contribute more to produce the output. In the arranged dataset, the first variable exhibits the highest correlation with the target, while the last demonstrates the lowest correlation. Tables \ref{tab2} and \ref{tab3} illustrate the six variables with the highest correlations with the targets for each training fold. Correlations close to 1 or -1 mean that the target highly correlates with the selected feature. 
\begin{table}
\centering
\caption{\label{tab2}Correlation scores of the first six important input features with the $mAb(t+1)$ target.}
\scalebox{0.9}{%
\begin{tabular}{c c c c c c c}
    \toprule
        Training &  $mAb(t)$ & $ECT$ & $Glutamine$ & $K^+$ & $EGN$ & $Osmolality$
        \tabularnewline
        \midrule 
        Fold-1 & $0.97$ &	$0.78$ & $0.77$ & $0.66$ &	$0.65$ & $0.53$ 	\tabularnewline
        Fold-2 & $0.97$ & $0.77$ &	$0.76$ & $0.67$ & $0.62$ &	$0.54$
        \tabularnewline
        Fold-3 & $0.97$ & $0.80$ &	$0.76$ & $0.65$ & $0.64$ &	$0.55$
        \tabularnewline
        Fold-4 & $0.97$ & $0.80$ &	$0.76$ & $0.67$ & $0.68$ &	$0.55$
        \tabularnewline
        Fold-5 & $0.97$ & $0.78$ &	$0.76$ & $0.64$ & $0.65$ &	$0.53$
        \tabularnewline
        \bottomrule
    \end{tabular}
    }
\end{table}
\begin{table}
\centering
\caption{\label{tab3}Correlation scores of the first six important input features with the $VCD(t+1)$ target}
\scalebox{0.9}{%
\begin{tabular}{c c c c c c c}
    \toprule
        Training &  $VCD(t)$ & $TCD$ & $Temperature$ & $pCO2$ & $HCO3^-$ & $Na^+$
        \tabularnewline
        \midrule 
        Fold-1 & $0.95$ &	$0.83$ & $-0.59$ & $-0.43$ &	$-0.41$ & $0.35$ 	\tabularnewline
        Fold-2 & $0.95$ & $0.83$ &	$-0.59$ & $-0.37$ & $-0.35$ &	$0.38$
        \tabularnewline
        Fold-3 & $0.95$ & $0.83$ &	$-0.63$ & $-0.39$ & $-0.36$ &	$0.33$
        \tabularnewline
        Fold-4 & $0.95$ & $0.83$ &	$-0.61$ & $-0.35$ & $-0.33$ &	$0.34$
        \tabularnewline
        Fold-5 & $0.95$ & $0.83$ &	$-0.61$ & $-0.36$ & $-0.35$ &	$0.35$
        \tabularnewline
        \bottomrule
    \end{tabular}
    }
\end{table}

In the initial experiment, the predictive model was trained using the first important variable, and the RMSE on the validation data was computed; then, the following essential variables were sequentially incorporated, and their respective RMSE on the validation data was computed. This process was repeated for all $23$ input variables, and the top $k$ features were considered the most critical variables for training the predictive model; here, $k$ represents the number of features resulting in the lowest RMSE on the validation fold. Table \ref{tab4} shows the validation RMSE for each fold, and Tables \ref{tab6} and \ref{tab5} show the test RMSE for each fold. However, the proposed model has one hyperparameter $\theta$, which had to be optimized. Due to the significant impact of the hyperparameter $\theta$ on the results, we performed a parameter tuning procedure with various values of $\theta = 0.1, 0.2, 0.3, ..., 0.7$ to identify the optimal $\theta$ for each training fold using another inner 5-fold cross-validation process. 
Then, the optimal $\theta$ between $0.1$ and $0.7$ for each training fold is selected, and the model is trained using optimal $\theta$. Finally, features in the top-k list of at least three folds were chosen as the best features for developing predictive models.

Table \ref{tab4} illustrates the number of input variables used in each fold to reach the best performance of the feature selection algorithm. According to the results, the first eight variables for VCD and the first seven for mAb with the highest correlation with the targets are the best selections that can lead to low validation RMSE scores for most validation folds. Results suggest that adding extra variables does not substantially enhance accuracy; instead, it elevates system complexity and learning time and could potentially diminish the predictive model's performance. Therefore, input variables of the $mAb$ prediction for each culture are illustrated in Eq.~\eqref{mab_feature_selection}:
\begin{equation}\label{mab_feature_selection}
    C'_i(t)=[mAb_i(t), ECT_i(t), Glutamine_i(t), K_i^+(t), EGN_i(t), Osmolality_i(t), ACV_i(t)]_{1\times7}
\end{equation}
with the target of $mAb_i(t+1)$. Similarly, input variables for the $VCD$ prediction are:\begin{equation}\label{vcd_feature_selection}
 \begin{split}
C''_i(t) = [&VCD_i(t), TCD_i(t), Temperature_i(t), Na_i^+(t), pCO2_i(t), HCO3_i^-(t), EGN_i(t), ECT_i(t)]_{1\times8}
 \end{split}
 \end{equation}
with the target of $VCD_i(t+1)$. It can be observed that the mAb concentration and VCD of the current day have a significant impact on the predictive performance of the following day. Regarding other process parameters, our feature selection technique accurately identifies the relevant parameters regarding their biochemical significance for cell growth and mAb concentration in cell culture bioreactors. For instance, glutamine is an essential supplement crucial for sustaining cell growth and product concentration \citep{shjo14, pera20}. Salts such as sodium ($Na^{+}$) and potassium ($K^{+}$) play a crucial role in various cellular processes, including transmembrane potential, nutrition, buffering, osmolality, and signal transduction \citep{riwu18}. These processes, in turn, affect cell growth and mAb titer. Meanwhile, high values of temperature, partial pressure of carbon dioxide ($pCO2$), and bicarbonate ($HCO3^{-}$) can negatively influence the cell growth, metabolism, and mAb productivity \citep{goma07, mabu15}, and they need to be controlled within acceptable ranges. Total cell density (TCD) also positively correlates with VCD in the cell culture process \citep{RN19}. In the literature, culture osmolality has consistently shown a positive effect on mAb productivity and titer \citep{alko21}. In the cell culture process, cell culture time (ECT) and the number of cell divisions (represented by the EGN feature) can significantly impact mAb concentration and overall cell growth. Antibody concentration might be low in the early phases of cell culture because of low cell densities. As the culture progresses, cells multiply, and the mAb concentration typically increases. However, when culture time is prolonged, cell growth might slow down due to overcrowding, nutrient depletion, or accumulation of inhibitory metabolites, decreasing mAb production and cell viability. Similarly, higher numbers of cell divisions initially result in increased cell numbers and mAb concentration. However, with continued divisions, cells might experience replicative senescence, decreasing viability, and mAb yield.

Using the selected feature subset in Eq.~\eqref{mab_feature_selection}, the training and testing sets for each experiment are defined as follows:
\begin{itemize}
    \item Training dataset: all $C'_i (t)$  where $i$ are cultures within the four training folds, and $1\le t\le 14$
    \item Testing dataset: all  $C'_i (t)$ where $i$ are cultures within the testing fold, and $1\le t\le 14$
    \item The target of dataset is $mAb_i(t+1)$.
\end{itemize}
The $VCD$ training and testing datasets are produced similarly to $mAb$. The predictive models aim to predict the future values for the $mAb$ concentration and $VCD$, and they are defined as follows:
\begin{equation}
    \widehat{mAb}_i(t+1)=\mathcal{F}(C'_i(t))
\end{equation} 
\begin{equation}
    \widehat{VCD}_i(t+1)=\mathcal{G}(C''_i(t)),
\end{equation} 
where $\mathcal{F}$  and $\mathcal{G}$ are the trained models and $\widehat{mAb}_i$ and $\widehat{VCD}_i$ shows the predicted values. The best prediction would be the prediction that minimizes the following equations:
\begin{equation}\label{eq23}
    \mathcal{L}=||\widehat{mAb}-mAb||^2
\end{equation}
\begin{equation}\label{eq24}
    \mathcal{L}=||\widehat{VCD}-VCD||^2,
\end{equation}
where $mAb$ and $VCD$ are actual values of cell cultures. The learning algorithms optimize parameters to minimize $\mathcal{L}$ in the above equations. The experiments in the next subsection utilize the selected features to train the predictive models for one-day-ahead mAb and VCD predictions and compare the various learning models in terms of accuracy, complexity, and learning time.
\begin{table}
\centering
\caption{\label{tab4}The best validation RMSE and top-K input features for the VCD and mAb prediction using 5-fold cross validation}
\begin{tabular}{c c c|c c}
    \toprule
        Val fold &  \multicolumn{2}{c}{$\widehat{VCD}_i(t+1)$} & \multicolumn{2}{c}{$\widehat{mAb}_i(t+1)$}  
        \tabularnewline
        \midrule 
         & Best Val RMSE &	Best top-K features & Best Val RMSE & Best top-K features
        \tabularnewline
        1 & $0.0372 \pm  0.0048$ & $8$ & $0.0431 \pm 0.0077$ & $6$
        \tabularnewline
        2 & $0.0410 \pm 0.0052$ & $8$ & $0.0432 \pm 0.0035$ & $11$
        \tabularnewline
        3 & $0.0392 \pm  0.0103$ & $8$ & $0.0427 \pm 0.0057$ & $5$
        \tabularnewline
        4 & $0.0415 \pm  0.0058$ & $9$ & $0.0435 \pm 0.0056$ & $6$
        \tabularnewline
        5 & $0.0409 \pm 0.0055$ & $9$ & $ 0.0395 \pm 0.0095$ & $8$
        \tabularnewline
        \bottomrule
    \end{tabular}
\end{table}
\begin{figure*}[t]
    \includegraphics[scale=0.28]{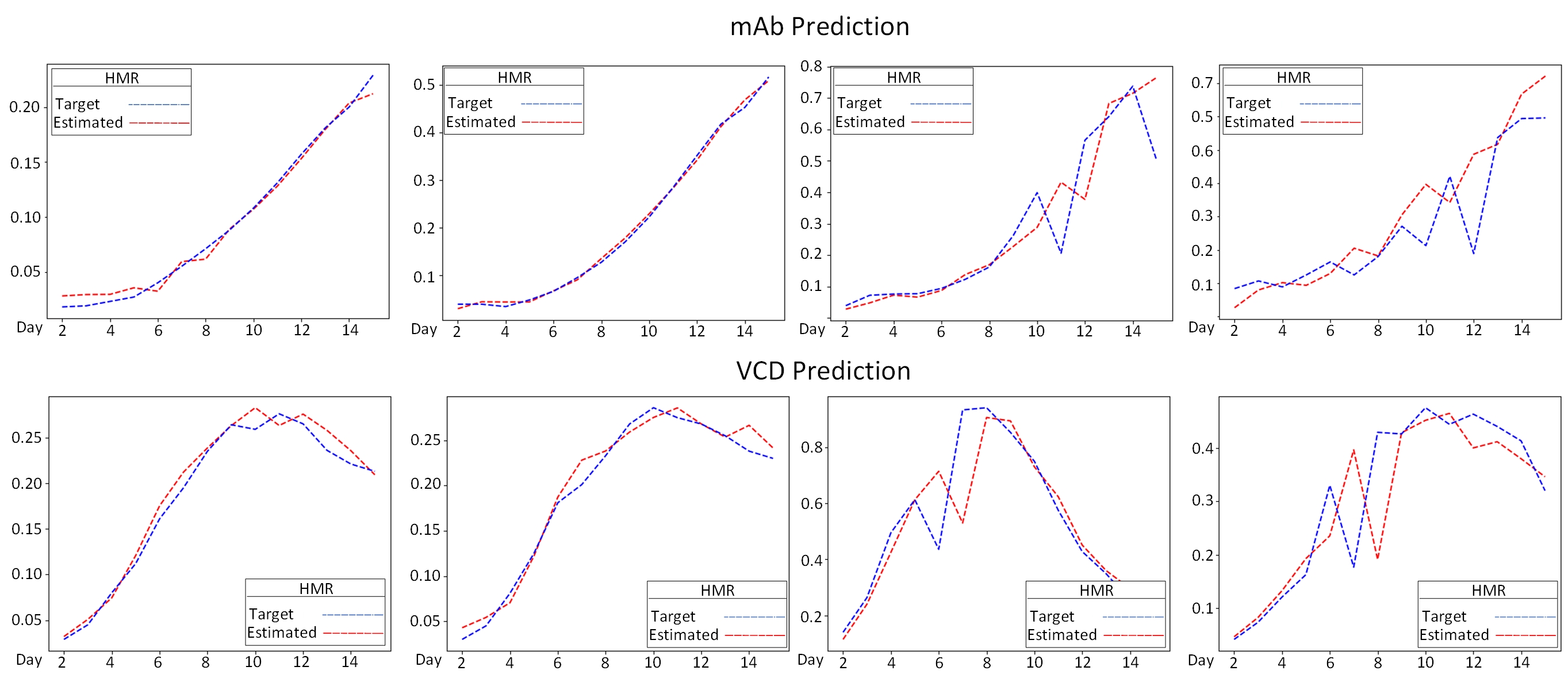}
    \caption{\label{Fig2}One-day-ahead prediction of the mAb concentration and VCD using HMR for representative bioreactors. The best predictions are illustrated in the two left figures, while the worst predictions are depicted in the two right figures.}
\end{figure*}
\subsection{Low dimensional scenario}\label{part_3}
In this section, experiments compare accuracy, complexity, and learning time between the competing predictive models in the low-dimensional scenario. Predictive models aim to predict one-day-ahead values of the mAb concentration and VCD. In this part, we employed the dataset with features selected in the previous part. 

To enhance the evaluation process, we initially optimized hyperparameters for each model. The optimization of hyperparameters involves employing a grid-searching method \citep{RN20} with 5-fold cross-validation on the training fold. In adherence to this approach, predictive models undergo training using various combinations of hyperparameters, and the set of hyperparameters yielding the lowest average RMSE on all validation folds is selected as the optimal configuration. After hyperparameter optimization and model training, the values of training and testing RMSE, the number of generated neurons in the first layer, training time, and tuning time for hyperparameter optimization are compared among different models. 
\begin{table}
\centering
\caption{\label{tab5} $mAb$ prediction RMSE and number of neurons in the first layer for predictive models using selected input features and 5-fold cross validation}
\begin{tabular}{c c c c}
    \toprule
         Predictive model &  ANFIS HL & FNN BP & \textbf{Proposed Model}
    \tabularnewline
    \midrule 
        $1^{th}$ layer size &  $128 \pm 0 $ & $128 \pm 0 $ & $32 \pm 6 $ 
        \tabularnewline
        Training time & $145.2$ Sec & $105.8$ Sec& $5.8$ Sec   
        \tabularnewline
        Tuning time &  $107.2$ Min & $84.9$ Min& $6.5$ Min
        \tabularnewline
        Train RMSE & $0.0303 \pm 0.0021 $ & $0.0561 \pm 0.0091 $ & $0.0392 \pm 0.0016 $  
        \tabularnewline
        Test RMSE & $0.0989\pm 0.0390 $ & $0.0701 \pm 0.0298 $ & $0.0455 \pm 0.0062 $ 
        \tabularnewline
        \bottomrule
    \end{tabular}
\end{table}
\begin{table}
\centering
\caption{\label{tab6} $VCD$ prediction RMSE and number of neurons in the first layer for predictive models using selected input features and 5-fold cross validation}
\begin{tabular}{c c c c}
    \toprule
         Predictive model &  ANFIS HL & FNN BP & \textbf{Proposed Model}
    \tabularnewline
    \midrule 
        $1^{th}$ layer size &  $256 \pm 0$ & $256 \pm 0$ & $133 \pm 2$ 
        \tabularnewline
        Training time & $605.3$ Sec & $91.6$ Sec& $44.4$ Sec   
        \tabularnewline
        Tuning time &  $230.7$ Min & $107.1$ Min& $6.0$ Min
        \tabularnewline
        Train RMSE & $0.0279 \pm 0.0019 $ & $0.0722 \pm 0.0151  $ & $0.0348 \pm   0.0017 $  
        \tabularnewline
        Test RMSE & $ 0.0680 \pm  0.0126 $ & $0.0755 \pm 0.0144 $ & $0.0409 \pm 0.0060 $ 
        \tabularnewline
        \bottomrule
    \end{tabular}
\end{table}

Table \ref{tab5} and \ref{tab6} illustrate the best performance of predictive models for one-day-ahead mAb and VCD prediction. The results suggest that, in these experiments, HMR produces significantly fewer neurons (i.e. hyperboxes equivalent to the fuzzy if-then rules in other methods) than the competing models to reach its best performance, signifying its lower complexity than other models. Additionally, the HMR model achieves lower test RMSE value, suggesting superior accuracy compared to other predictive models. Furthermore, the proposed model exhibits significantly higher learning speeds than other alternative models in training and tuning processes. Moreover, the training and testing RMSE scores in the proposed model are close to each other, implying its good generalization. Notably, the testing accuracy of all predictive models using the selected subset of features is significantly higher than when using all 23 input features, as depicted in Table \ref{tab1}.    

According to the above outcomes, the proposed HMR model significantly outperforms the conventional ANFIS models. The proposed model offers lower complexity, a higher learning rate, and accuracy with no curse of dimensionality limitation. Results also confirm that the proposed model performs better than conventional ANFIS in both low and high-dimensional problems. Next, we assess the proposed model for one-day-ahead and two-day-ahead mAb and VCD prediction.

\subsection{Analyses of HMR in the process performance prediction}\label{part_4}
This section involves comprehensive experiments to assess the proposed HMR model's ability to predict mAb and VCD one and two days ahead. In this section, the proposed model employs hyperparameters and features selected in previous parts. We trained the model using 5-fold cross-validation for a better assessment and computed the RMSE of all 106 bioreactors during their respective 14 days. 

\begin{figure*}
    \includegraphics[scale=0.4]{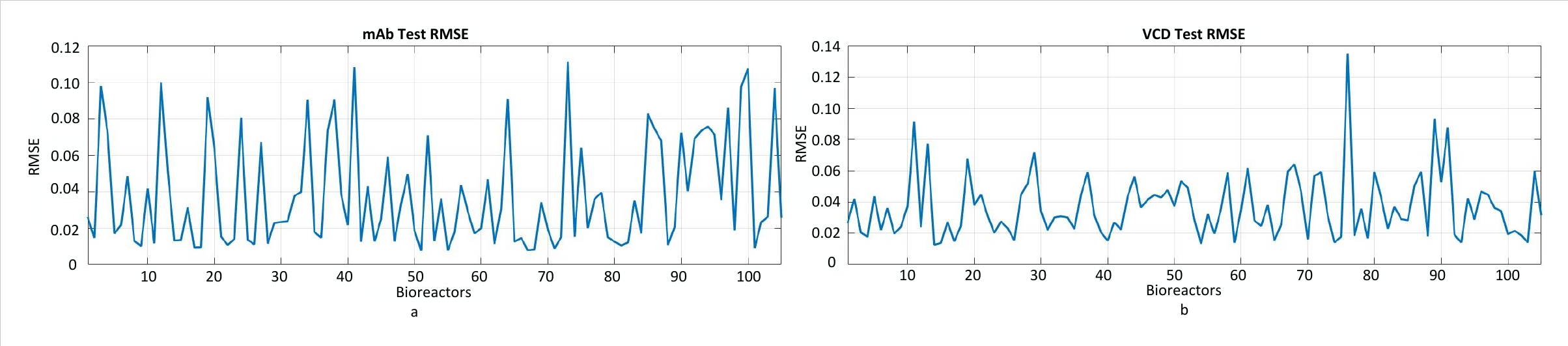}
    \caption{\label{Fig3} Testing RMSE of a) mAb and b) VCD one-day-ahead prediction for 106 bioreactors using HMR.}
\end{figure*}

Fig. \ref{Fig2} shows the best and the worst mAb and VCD predictions using the HMR model. In addition, Fig. \ref{Fig3} shows the testing RMSE scores of $106$ bioreactors across their respective 14 culture days. Results demonstrate that the proposed model can track the target accurately. However, the performance of predictors is reduced for cultures experiencing sudden increases or decreases in the mAb concentration and VCD on the next culture day compared to the current day, as depicted in the worst-case scenarios in Fig. \ref{Fig2}. This inconsistency might arise from operator interventions in the cell culture process, such as adding specific nutrients (e.g., glucose) that are not included in the reduced set of features. Additionally, Table \ref{tab7} illustrates that the proposed model can more accurately predict the first five days of the cell culture process, while the predictive performance remains consistent for the subsequent culture days. However, the HMR model exhibits more significant variation in predicting mAb concentrations among 106 cultures from the $6^{th}$ culture day compared to the initial five days. This variation is evident through higher standard deviation values in Table \ref{tab7}. 

In addition, another HMR model was trained to predict two-day-ahead mAb and VCD values. In the following experiments, we employed $C_i^3$  and $C_i^4$ to train the predictive model for two-day-ahead prediction using 5-fold cross-validation. Therefore, the mAb and VCD input variables are defined as follows:
\begin{equation}
    C^3_i(t)=[mAb_i(t), ECT_i(t), Glutamine_i(t), K_i^+(t),EGN_i(t), Osmolality_i(t), ACV_i(t), mAb_i(t+1)]_{1\times8}
\end{equation}
with the target of $mAb_i(t+2)$. Similarly, input variables for $VCD$ prediction are:
\begin{small}
 \begin{equation}
C^4_i(t) = [VCD_i(t), TCD_i(t), Temperature_i(t), Na_i^+(t), pCO2_i(t), HCO3_i^-(t), EGN_i(t), ECT_i(t),VCD_i(t+1)]_{1\times9}
 \end{equation}
 \end{small}
with the target of $VCD_i(t+2)$. Using these definitions, the dataset for each experiment is presented as follows:
\begin{itemize}
    \item Training dataset: all $C^3_i(t)$  where $i$ are cultures within the four training folds, and $1\le t\le 13$
    \item Testing dataset: all  $C^3_i(t)$ where $i$ are cultures within the testing fold, and $1\le t\le 13$
    \item The target of dataset is $mAb_i(t+2)$.
\end{itemize}

The $VCD$ training and testing datasets are produced similarly to $mAb$. The following models are trained using the above equations:
\begin{equation}
    \widehat{mAb}_i(t+2)=\mathcal{F}(C^3_i(t))
\end{equation} 
\begin{equation}
    \widehat{VCD}_i(t+2)=\mathcal{G}(C^4_i(t)),
\end{equation} 
\begin{table}[!ht]
\centering
\caption{\label{tab7} Mean testing RMSE of each day across $106$ bioreactors for $VCD$ and $mAb$}
\scalebox{1}{%
\begin{tabular}{c c c}
    \toprule
         Predicted Day &  mAb Test RMSE & VCD Test RMSE
    \tabularnewline
    \midrule 
        Day 2 &  $0.0125 \pm   0.0179$ & $0.0098 \pm 0.0099$
        \tabularnewline
        Day 3  & $0.0102 \pm  0.0101$  & $0.0152 \pm 0.0172$ 
        \tabularnewline
        Day 4 &  $0.0130 \pm 0.0159$  & $0.0255 \pm 0.0299$ 
        \tabularnewline
        Day 5  & $0.0143 \pm 0.0116$  & $0.0247 \pm 0.0252$ 
        \tabularnewline
        Day 6  & $0.0210 \pm 0.0220$  & $0.0405 \pm 0.0393$ 
        \tabularnewline
        Day 7  & $0.0241 \pm 0.0245$  & $0.0419 \pm 0.0569$ 
        \tabularnewline
        Day 8  & $0.0248 \pm 0.0300$  & $0.0306 \pm  0.0343$ 
        \tabularnewline
        Day 9  & $0.0298 \pm 0.0332$  & $0.0287 \pm 0.0274$ 
        \tabularnewline
        Day 10  & $0.0356 \pm 0.0437$  & $0.0278 \pm 0.0229$ 
        \tabularnewline
        Day 11  & $0.0387 \pm  0.0499$  & $0.0308 \pm 0.0274$ 
        \tabularnewline
        Day 12  & $0.0381 \pm  0.0584$  & $0.0333 \pm 0.0323$ 
        \tabularnewline
        Day 13  & $0.0434 \pm 0.0558$  & $0.0248 \pm  0.0235$ 
        \tabularnewline
        Day 14  & $0.0365 \pm 0.0504$  & $0.0210 \pm 0.0216$ 
        \tabularnewline
        Day 15  & $0.0343 \pm 0.0391$  &  $0.0231 \pm 0.0212$  
        \tabularnewline
        \bottomrule
    \end{tabular}
           }
\end{table}
The optimization algorithms aim to minimize the overall predicted values given in Eqs.~\eqref{eq23} and \eqref{eq24} to determine optimal parameters of the predictive models. Additionally, it is essential to note that the second model requires the actual values of $mAb_i(t+1)$ and $VCD_i(t+1)$ as input features in the testing phase, which is impossible to obtain in reality. Consequently, during the testing phase of the trained model, we utilized the $\widehat{mAb}_i(t+1)$ and $\widehat{VCD}_i(t+1)$ instead of their actual values. Hence, the revised prediction equations for the model are presented as follows:
\begin{equation}
    \widehat{mAb}_i(t+2)=\mathcal{F}(C'_i(t), \widehat{mAb}_i(t+1))
\end{equation} 
\begin{equation}
    \widehat{VCD}_i(t+2)=\mathcal{G}(C''_i(t), \widehat{VCD}_i(t+1)),
\end{equation} 
\begin{table}
\centering
\caption{\label{tab8} $VCD$ and $mAb$ two-day-ahead prediction testing RMSE and number of neurons in the first layer for the Proposed Model}
\begin{tabular}{c c c }
    \toprule
         Predictive model & $ \widehat{mAb}_i(t+2)$ & $ \widehat{VCD}_i(t+2)$
    \tabularnewline
    \midrule 
        $1^{th}$ layer size &  $25 \pm 1$ & $129 \pm 6$ 
        \tabularnewline
        Training time & $3.7$ Sec & $40.7$ Sec  
        \tabularnewline
        Tuning time &  $2.5$ Min & $4.2$ Min
        \tabularnewline
        Train RMSE & $0.0342 \pm 0.0012$ & $0.0348 \pm 0.0014$  
        \tabularnewline
        Test RMSE & $ 0.0390 \pm 0.0051  $ & $0.0422 \pm 0.0075$ 
        \tabularnewline
        \bottomrule
    \end{tabular}
\end{table}
This model is utilizable in reality.
Therefore, the first model predicts the next day's VCD and mAb values and another predictive model uses the prediction to estimate the VCD and mAb values for the next two days. Table \ref{tab8} shows the proposed model performance for two-day-ahead prediction of the mAb concentration and VCD values. According to the results, the proposed model effectively generates critical basis functions, reducing complexity and accelerating the learning rate.

\section{Conclusion}
This paper introduced the HMR model as a novel approach to predicting bioprocess performance in monoclonal antibody production. Our findings demonstrate that the HMR model effectively addresses the challenges posed by high-dimensional time-series data, significantly improving predictive accuracy and learning speed compared to traditional statistical methods and other machine learning models.

The HMR model's unique ability to partition the input space using hyperboxes allows for enhanced interpretability and robustness, particularly in the face of uncertainty common in bioprocess data. The inclusion of hyperboxes offers advantages, such as learning data in a single-pass process and dynamic basis functions genertion. In addition, by employing a local linear regressor within each hyperbox, we achieved greater accuracy while reducing computational complexity, making the model suitable for real-time applications in biopharmaceutical production.

The effectiveness of the proposed method was assessed on the real-world mAb production problem with different scenarios. In our experimental study, we trained the predictive models in both high and low dimensions for better evaluation. According to the empirical outcomes, the HMR can accurately estimate the output target in both scenarios. In addition, due to dynamic basis functions generation, the HMR only generates the critical basis functions, which leads to a less complex structure and avoids the curse of dimensionality problems. 

However, the HMR is sensitive to the expansion coefficient $\theta$. In other words, as we decrease the $\theta$, the HMR generates more hyperboxes, which increases complexity. On the other hand, increasing the expansion coefficient $\theta$ might affect the accuracy of the predictive models. In this paper, we employed grid-search for hyperparameter optimization but in future works, we aim to develop the learning procedure to make it a $\theta$-independent approach. Additionally, to improve the model's performance in uncertain conditions, the multi-stream learning approach proposed in \citep{yu2022real} can be considered in the future.

Methods previously developed for robust hyperbox clustering systems exploiting various ensembling and method-independent statistical learning approaches \citep{ga02a,ga04} will be explored and adopted to the regression settings considered in this paper. When applying the proposed method to predict the process performance in mAb production, the current values of process parameters were employed as input features to forecast the following day's culture. However, another model-building approach that incorporates actual values up to the current day within each process parameter as input features for early predictions regarding subsequent culture days is feasible and should also be considered in future studies.

\section*{Acknowledgment}
This research was supported under the Australian Research Council's Industrial Transformation Research Program (ITRP) funding scheme (project number IH210100051). The ARC Digital Bioprocess Development Hub is a collaboration between The University of Melbourne, University of Technology Sydney, RMIT University, CSL Innovation Pty Ltd, Cytiva (Global Life Science Solutions Australia Pty Ltd), and Patheon Biologics Australia Pty Ltd.

\bibliographystyle{elsarticle-num-names}
\bibliography{References.bib}

\end{document}